# Automatic identification of fossils and abiotic grains during carbonate microfacies analysis using deep convolutional neural networks

Xiaokang Liu[1], Haijun Song[1]*

[1]State Key Laboratory of Biogeology and Environmental Geology, School of Earth Sciences, China University of Geosciences, Wuhan 430074, China

*Corresponding author. Email: haijunsong@cug.edu.cn

## Abstract

Petrographic analysis based on microfacies identification in thin sections is widely used in sedimentary environment interpretation and paleoecological reconstruction. Fossil recognition from microfacies is an essential procedure for petrographers to complete this task. Distinguishing the morphological and microstructural diversity of skeletal fragments requires extensive prior knowledge of fossil morphotypes in microfacies and long training sessions under the microscope. This requirement engenders certain challenges for sedimentologists and paleontologists, especially novices. However, a machine classifier can help address this challenge. In this study, we collected a microfacies image dataset comprising both public data from 1133 references and our own materials (including a total of 30,815 images of 22 fossil and abiotic grain groups). We employed a high-performance workstation to implement four classic deep convolutional neural networks, which abstracthave proven to be highly efficient in computer vision. Our framework uses a transfer learning technique, which reuses the pre-trained parameters that are trained on a larger ImageNet dataset as initialization for the network to achieve high accuracy with low computing costs. We obtained up to 95% of the top one and 99% of the top three test accuracies in the Inception ResNet v2 architecture. The machine classifier exhibited 0.99 precision on minerals such as dolomite and pyrite. Although it had some difficulty on samples having similar morphologies, such as bivalve, brachiopod, and ostracod, it nevertheless obtained 0.88 precision. Our machine learning framework demonstrates high accuracy with reproducibility and bias avoidance that is comparable to those of human classifiers. Its application can thus eliminate much of the tedious, manually intensive efforts by human experts conducting routine identification.

## 1. Introduction

Most petrographic analysis depends on microscopic observation of thin sections. This procedure is typically labor intensive and requires substantial prior knowledge (Wilson, 1975; Flügel, 2010). Most students who examine sedimentary rocks struggle to identify the microfacies, such as the standard microfacies types in Flügel (2010). The most important procedure is to distinguish the morphological and microstructural diversity of skeletal grains. Owing to differences in examiners' abilities and subjective perceptions, identification of fossils and sedimentary structures in thin sections remains challenging for petrographers, especially novices. Meanwhile, machine learning algorithms have recently demonstrated high accuracy in image recognition tasks in both computer vision (LeCun et al., 2015; He et al., 2016) and paleontology (Hsiang et al., 2019; Bourel et al., 2020). Hence, we believe that machine learning algorithms can help petrographers to identify fossils during microfacies analysis with reliability and objectivity.

Machine learning has advanced considerably over the last decade (MacLeod et al., 2010; LeCun et al., 2015). This progress is partly attributed to the fact that the use of graphical processing units (GPUs) has significantly increased the speed of training and classification operations in convolutional neural networks (CNNs). The CNN architectures also benefit from GPU acceleration. Hence, scientists have begun to develop deep convolutional neural networks (DCNNs). The ImageNet Large Scale Visual Recognition Challenge (ILSVRC), which contains 1000 categories and over 1.2 million images, has achieved a top-5 error rate of 2.3% on the validation set in 2017 (Russakovsky et al., 2015). This achievement almost surpasses human identification accuracy. Automatic identification based on machining learning in the fields of sedimentology and paleontology has also been proposed, such as the classification of sedimentary rocks (Shu et al., 2017; Baraboshkin et al., 2020), petrography analysis (Izadi et al., 2017; Shu et al., 2018; Duarte-Coronado et al., 2019; Pires de Lima et al., 2019a; 2019b; 2020a), and biotic identification, including planktonic foraminifera (Hsiang et al., 2019; Mitra et al., 2019), fusulinids (Pires de Lima et al., 2020b), pollen grains (Marcos et al., 2015; Bourel et al., 2020), coccoliths (Beaufort and Dollfus, 2004), insects (Larios et al., 2008; Rodner et al., 2015; Valan et al., 2019), benthic invertebrates (Lytle et al., 2010), fish bones and teeth (Hou et al., 2020), and diatoms (Urbankova et al., 2016).

The early application of neural networks in petrography used composite well data to enable petrophysical identification (Baldwin et al., 1990; Chang et al., 2002; Marmo et al., 2005). With the rise of artificial intelligence and computer vision in the last decade, petrographic analysis studies have diversified into various areas, such as the sorting of particles (Shu et al., 2018), rock classification (Cheng and Guo, 2017; Shu et al., 2017), and microfacies classification (Pires de Lima et al., 2020a). Budennyy et al. (2017) employed machine learning to evaluate the properties of structural objects in thin sections, such as grains, cement, voids, and cleavage. Their models achieved up to 80% accuracy and proved a way to conduct automatic quantitative and qualitative analysis of thin sections by applying image processing and statistical learning methods. Pires de Lima et al. (2020a) implemented a relatively small number of labeled thin sections (with five classes of rocks from 98 thin sections) used in the fine-tuning method (explained below) and achieved error levels lower than 5% for the classification of microfacies from the same dataset.

Despite the above efforts, most studies using machine learning employed only local or personal samples materials for convenience. It remains challenging to apply these models for prediction using samples from other regions. For instance, the fossil assemblages of different geological periods may be quite different (Flügel, 2010). Furthermore, some studies used only a limited number of rock types, microfacies classes, or fossil groups, resulting in insufficient actual classification.

Meanwhile, DCNNs require massive training datasets and can thus produce accurate identifications. In this study, we focused on identifying fossil fragments from microfacies. We collected 22 of the most common types of fossil and mineral groups as well as sedimentary structures from both the published literature and our own collection. More than 30,000 images were used. We implemented four classic DCNNs that have a high test accuracy.

## 2. Materials and data

In this study, we used both image data collected from publicly available literature and our materials from the Permian and Triassic in South China. A total of 21,839 images (almost all of them from thin sections) were collected from 1133 references (An appendix file provided for original references) from the Phanerozoic Eon; A total of 8976 images were from our samples in South China (Song et al., 2009a, 2009b; Song et al., 2015; Dai et al., 2018; Jia and Song, 2018; Liu et al., 2020, plus additional unpublished materials). The majority of the public data was collected from thin sections reported in the literature. We use Adobe Acrobat DC to crop the separate fossil and abiotic grains from published images. We do not strictly limit the size and resolution of the images, as long as its quality is sufficient for identification. The database ultimately contains 30,815 images of 18 fossil groups as well as four minerals or sedimentary structures: algae, bivalve, brachiopod, bryozoan, calcimicrobe, calcisphere, calpionellid, cephalopod, coral, echinoderm, foraminifer, gastropod, ostracod, radiolarian, sponge, stromatolite, stromatoporoid, *Tubiphytes*, dolomite, oncolite, ooid, and pyrite (Fig. 1). We randomly divided all data into three categories: training set (80% of data), validation set (15% of data), and test set (5% of data). Details are provided in Table 1. The training set was used for model fitting, the validation set was used to tune the model hyperparameters and initially evaluate the model's ability, and the test set was used to evaluate the generalization ability of the final model.

# 3. Methods

All analysis codes were run on a Dell Precision 7920 Workstation desktop on Windows 10 Professional, including two Intel Xeon Silver 4216 Processors, 128 GB of RAM, and two NVIDIA GeForce GTX 2080Ti GPUs (11 GB for each GPU). The versions of Python and TensorFlow used were 3.6.5 and 1.13.1, respectively. The corresponding versions of the NVIDIA CUDA and cuDNN were 10.0 and 7.4, respectively. The algorithms for the DCNN analyses and model weights are available on GitHub (https://github.com/XiaokangLiuCUG/microfacies_analysis_with_dcnn).

## 3.1 Convolutional neural network

The concept of the artificial neural network is inspired by the biological neural networks that constitute animal brains (Hubel and Wiesel, 1962). Fig. 2A depicts a classic artificial neural network, Visual Geometry Group (VGG)-16 network (Simonyan and Zisserman, 2014). A neuron in an idle state accumulates all the signals it has received until it reaches a certain activation threshold (by using the Heaviside step function as activation function) (McCulloch and Pitts, 1943), which is also known as a perceptron. The basic functioning of a signal neuron or perceptron is shown in Fig. 2E and can be expressed as follows:

$$Y = \varphi(\sum_{i=1}^{n} w_i x_i + b),$$

where $x_i$ is the input value, $Y$ denotes the output result, $b$ represents the bias, $w_i$ is the weight, n is the number of the inputs, and $\varphi$ is the activation function (generally a nonlinear activation function, such as ReLU, sigmoid, and tanh). Further, $Y$ can be the input of the next neural or the final output of the neural network and eventually achieves classification or regression objectives.

The convolutional neural network (CNN), a class of deep neural networks, is designed and classically used for image recognition, such as image classification and tagging, object detection, and face detection and recognition (LeCun et al., 2015). CNNs are regularized versions of multilayer perceptrons (composed of the convolutional layer(s) and some other layers), as shown in Fig. 2A. The core concept of the CNN is the convolution operation; that is, each image is composed of a matrix of pixel values (usually with one or three channels represented in a grayscale or RGB model). Then, the sub-portion of the input matrix (the 5 × 5 matrix in Fig. 2B) can be placed on the convolution with the filter set (the 2 × 2 yellow matrix in Fig. 2B). Scrolling of each filter along the matrix computes the inner product of the same filter and input, and each output of the matrix is a feature map (or activation map, the 4 × 4 matrix in Fig. 2B, also shown in Fig. 3). Each feature map is followed by a nonlinear activation function to allow neural networks to learn complex decision boundaries (commonly used ReLU, $f(x) = \max(0; x)$, where x is the input data, $f(x)$ is output.) (Nair and Hinton, 2010). After the convolution and activation operations, a pooling layer (Fig. 2 C) is used to merge semantically similar features into a single feature (i.e., downsampling). The end of the CNN is commonly attached to the fully connected layer (Fig. 2D). That is, each neuron in one layer is connected to each neuron in the next layer. The output of the final fully connected layer is the expected output class. These processes, from input images to final output classifications, are called forward propagation (calculation of the network errors). For predicted images, a softmax layer (green rectangle in Fig. 2D) is added, which is used to generate the true probability vector.

Backpropagation is a learning algorithm for adjusting the weights during the training process (Wlodarczak, 2019). The weights are usually set randomly when the CNNs are initialized. After each training iteration, the error or loss at the output of the multilayer perceptron is calculated and propagated back through the network to adjust the weights and ensure error minimization in the next iteration. An appropriate optimizer algorithm provides a correct and effective direction to increase the performance of the neural networks. Commonly used optimizers are Adam, RMSprop, and SGD. An iteration is not complete until both forward propagation and backpropagation are completed. Theoretically, an epoch is completed when each image from the training dataset is fit for both forward propagation and backpropagation.

## 3.2 Transfer learning

Deep convolutional neural networks are accompanied by numerous training parameters. Taking the VGG-16 architecture as an example, there are 138 million parameters trained on the ImageNet dataset with 1.2 million labelled images, including 1000 classes (Deng et al., 2009; Simonyan and Zisserman, 2014; Russakovsky et al., 2015). However, in practice, we usually do not have enough data to newly train such complex DCNN architectures. Therefore, transfer learning is a high-efficiency approach to compensate for dataset insufficiency. We take DCNN's parameters that have already been trained in a larger dataset (such as ImageNet) and make appropriate changes to fit in our smaller dataset. Some layers are frozen (such as Conv1 to Conv5 layers in Fig. 2A, and its parameters will be untrainable) as a feature extractor. Additionally, we used a few trainable layers (fc6–fc8 layers in Fig. 2A) on the top of the frozen layers as the classifier. Another alternative method is fine-tuning, which unfreezes all the layers (or most of the layers) using the pre-trained parameters as initial values and re-training them on our dataset with a very low learning rate. Transfer learning can achieve high accuracy on a small dataset while reducing both the demand for training data and the training time (Tan et al., 2018).

### 3.3 Data augmentation

Data augmentation is an image pre-processing procedure and a means to "enlarge" the dataset. Data augmentation is effective in preventing overfitting (that is, when there is high accuracy on the training dataset, but much lower accuracy on the validation or test dataset) with a good generalization ability (Wong et al., 2016; Xu et al., 2016). When we use a small dataset to train the DCNNs, we can implement some pre-processing methods to adjust one image into several images to "deceive" the neural network, such as by (1) randomly horizontally and vertically flipping the images, (2) randomly rotating the images, (3) resizing the scale of the images, (4) randomly cropping the images, (5) transforming the pixel matrices (such as by subtracting the mean values), and (6) adjusting the color space of the images (such as brightness and contrast). VGG and ResNet use methods 1, 3, 4, and 5 (Simonyan and Zisserman, 2014; He et al., 2016). Inception uses methods 1, 3, 4, 5, and 6 (Szegedy et al., 2015).

### 3.4 Evaluation metrics

Evaluation metrics (Fig. 4) is a technique for visualizing, organizing, and selecting classifiers based on artificial neural network performance (Fawcett, 2006). It covers several metrics, such as true positive (TP), false positive (FP), true negative (TN), and false negative (FN) while considering one of the labels the positive label (which is usually an interest class). Several common metrics are derived from the confusion matrix: accuracy (the ratio of correctly predicted labels to the total observations); precision (the ratio of correctly predicted positive labels to the total predicted positive observations); recall (the ratio of correctly predicted positive labels to all observations in the actual class; and the F1 score (a comprehensive index that is the harmonic mean of the precision and recall). This score considers both false positives and false negatives (Sarkar et al., 2018).

### 3.5 Training architectures

In this study, we trained four classical DCNNs: VGG-16, ResNet v1-152, Inception v4, and Inception ResNet v2 (Fig. 5). These four DCNN architectures were trained on the ImageNet dataset (Russakovsky et al., 2015), and they performed with high accuracy. Considering the computational limitations of the hardware, we randomly fed the DCNNs with an appropriate batch size for each iteration. We ran each DCNN architecture several times to adjust the hyper-parameters, which could not be trained by the networks. All networks ran from 40 to 60 epochs. The training and validating accuracy were the output of one batch size during the training procedure, and the test accuracy was the average result of all images in the test dataset from each of the two training epochs.

The VGG-16 contains approximately 138 million trainable parameters with 16 layers (Simonyan and Zisserman, 2014). The size of the convolutional kernels used in the VGG-16 architecture is 3 × 3. The size of the maxpool kernels are of size 2 × 2 with a stride of two; The input image size is 224 × 224 × 3. The ResNet v1-152 contains approximately 60 million trainable parameters and 152 layers (Fig. 5A). It uses the method of residual modules and batch normalization techniques (He et al., 2016). Each residual block stacks up its input (identity mapping or skip connection) and output (residual mapping) to be the input of the next layer. The identity mappings can prevent the vanishing gradient issue from building even deeper layers. The input image size is the same as the VGG-16. The Inception architecture is called GoogLeNet. The Inception v4 architecture is composed of 148 layers and approximately 43 million trainable parameters (Fig. 5B). The GoogLeNet group devised a new notion known as "blocks of inception" (network in the network) (Lin et al., 2013; Szegedy et al., 2015), whereby it embeds a multiscale feature extractor, such as $1 \times 1$, $3 \times 3$, and $5 \times 5$ convolutions, within the same module of the network. The input size for this architecture is $299 \times 299 \times 3$. The Inception ResNet v2 architecture is composed of 164 layers and approximately 55 million trainable parameters (Fig. 5C). This network is inspired by ResNet and is a hybridization of Inception and ResNet architectures, and uses residual connections as an alternative to concatenation filters (Szegedy et al., 2016). The input size for this architecture is consistent with Inception v4.

## 4. Results

**4.1 VGG-16**

The best performance of VGG-16 reached 0.91 for the top one test accuracies and 0.98 for the top three test accuracies (analysis number #1 in Table 2). The corresponding minimum training and validation losses were 0.12 and 0.52, respectively (Fig. 6A). In the figure, the losses decline with increased fluctuations, and the accuracy slowly increases. A performance difference occurs between the models trained anew and the fine-tuned models. The former shows 0.6 test accuracy, whereas the latter has 0.91 test accuracy. There is a significant bottleneck or barrier of the model that is trained from the beginning. The validation loss starts to increase when the training procedure reaches eight epochs, whereas the training loss seems to decrease effectively. In this situation, with massive parameters, the model is more likely to overfit on the training dataset, which may explain the marked fluctuations in the validation loss of VGG-16 in Fig. 6C. We thus attempted dropout regularization (ranging from 0 to 1.0, where "0" represents probabilistically removing or "dropping out" all the inputs to the next layer and "1" denotes no dropping out) to solve the overfitting (Table 2). The architecture is barely improved when the dropout equals 0.5, and it reaches the optimal model when the dropout equals 0.8. The image pre-processing without crop training images (including treatments (1), (3), and (5) in section 3.3) improves approximately 3% of the test accuracy. VGG-16 identifies 0.87 of the average accuracy in the validation and test datasets. Among them, the prediction precision for calpionellid, radiolarian, pyrite, and dolomite were up to 0.94, whereas it is more difficult to identify algae (0.69) and bivalves (0.79). The averages of the recall and F1 score were $0.87\pm0.07$ and $0.87\pm0.06$ respectively. The former ranges from 0.72 to 1, and the latter ranges from 0.71 to 0.99 (see the Supplementary Table for details).

**4.2 ResNet v1-152**

The optimal performance of ResNet v1-152 recorded 0.94 of the top one test accuracies and 0.99 of the top three test accuracies (analysis number #7 in Table 2), which was significantly higher than those of VGG-16. The minimum training and validation losses were 0.29 and 0.36, respectively (Fig. 6A). The model that trained from the beginning also represented overfitting when training stepped up to 20 epochs, and it showed a final test accuracy of 0.68. ResNet v1-152 also demonstrated accurate performance (0.91 accuracy, analysis number #8 in Table 2) when the training images were randomly cropped. The dropout function in the last layer had only a slight contribution to model optimization. The average accuracy of the validation and test datasets for analysis number #7 was 0.89, and the precision ranged from 0.79 (bivalve) to 1.00 (pyrite). The average F1 score of these 22 classes was $0.89 \pm 0.05$. The highest recall score was for pyrite (1.00), while the lowest for brachiopod obtained 0.79 (also in Supplementary Table).

**4.3 Inception v4**

Inception v4 exhibited 0.94 of the top one test accuracies and 0.99 of the top three test accuracies (analysis number #13 in Table 2). The corresponding minimum training and validation losses were 0.19 and 0.28, respectively (Fig. 6A). Inception v4 also implements batch normalization with good adaptability and quick convergence. The default pre-processing method of the inception architecture uses randomly cropped images, which include a 0.05:1 proportion of the original images for network training. The result was approximately 5% lower than those of the no-crop models (including treatments (1), (3), (5), and (6) in section 3.3). This architecture demonstrated a lower training loss compared with the other three architectures (Fig. 6A). However, the validation loss was inconspicuous, which may be attributed to overfitting or the use of fewer parameters. The final average precision on the validation and test datasets was $0.91 \pm 0.04$. Among them, algae (0.83) and oncolite (0.84) had relatively low precision, whereas radiolarian (0.99) and pyrite (0.99) had high precision. The algae bivalve and sponge contained low recall scores, that is, low sensitivity.

The averages of recall and F1 score are 0.92±0.05 and 0.91±0.04 respectively.

**4.4 Inception ResNet v2**

The Inception ResNet v2 architecture obtained the highest top one test accuracy (0.95) and top three test accuracies (0.99) in all four DCNNs. In this architecture, the minimum training and validation losses were 0.39 and 0.40, respectively. By combining the advantages of Inception and ResNet networks, Inception ResNet v2 exhibited an effective path for model convergence (Fig. 6D), which was the fastest one to reach the 1.00 validation accuracy (as shown in Fig. 7). In addition, this architecture achieved the highest accuracy in the cropped training images with 0.93 test accuracy (analysis number #22 in Table 2). The average precision and F1 score for all classes on the validation and test datasets were 0.93 ± 0.04. The lowest precision was 0.88, which was for ostracods (Fig. 8). Dolomite had a precision of 1.00. Classes of bivalve, brachiopod, and oncolite demonstrated 0.87 F1 scores.

# 5. Discussion

**5.1 Performance evaluation**

We trained four classic DCNNs for image recognition and ran each architecture several times to fine-tune the hyper-parameters. These four architectures showed different performances in fossil classification from the thin sections. Among them, VGG-16 exhibited inferior evaluation performance (0.91 of test accuracy), and Inception ResNet v2 obtained the best performance (0.95 of test accuracy, model weights are available on

https://github.com/XiaokangLiuCUG/microfacies_analysis_with_dcnn). The VGG-16 is more notably affected by overfitting, and its parameters are several times larger than those of other networks. We also attempted to use convolutional layers instead of the fully connected layers (i.e., fc6, cf7, and fc8, which included 86% of all the parameters) in the posterior three layers. This method partially mitigated the effect of overfitting, and the training and validation loss curves decreased more stably during the 40 training epochs. However, the test accuracy was scarcely improved (0.88).

In addition, we implemented dropout regularization and data augmentation to reduce overfitting on VGG-16, and its effectiveness was distinct. L1/L2 regularization did not improve the validation accuracy on the planktonic foraminifera dataset (Hsiang et al., 2019). A similar overfitting situation occurred on other datasets (Hsiang et al., 2019; Kaya et al., 2019; Mott, 2019). Simonyan and Zisserman (2014) suggested using VGG-16 pre-trained parameters on ILSVRC as feature extractors on other smaller datasets because training large models anew may not be feasible owing to overfitting. Furthermore, in theoretical terms, a deeper network helps the model to extract more features; however, it is more prone to "explode" the gradient (causing it to vanish), resulting in a network's inability or failure to converge (He et al., 2016; Zaccone et al., 2017; Hanin, 2018). Hence, the VGG architecture should use it prudently, especially when the dataset is not large enough for this architecture (Chatfield et al., 2014).

Data augmentation is the most common way to improve the generalization ability of networks in computer vision. Data augmentation is essentially a type of regularization. Using appropriate data augmentation methods can improve the generalization ability and accuracy of the model; nevertheless, it will require a longer training time and thus increased costs. In our dataset, the data augmentation technology led to lower accuracies in all architectures in 40–60 training epochs. However, by using a combination of several regularizations, such as dropout, L1/L2 regularization, and early stopping (a technique that monitor the architecture's performance and stop training automatically when the model stopped improving), it could easily lead to underfitting (Zheng et al., 2016; Czyzewski, 2020). With excessive "dropout," the model may have difficulty obtaining sufficient information to continue model optimization, leading to learning bottlenecks. This scenario occurred on the VGG-16 and Inception v4 architectures, in which the validation error barely decreased after 10–15 epochs, whereas Inception ResNet v2 showed greater efficiency (red curves in Fig. 7).

Although underfitting can be compensated by increasing the training epochs or by employing larger batch sizes, the result may still be insignificant (Hsiang et al., 2019). Hence, in terms of the relatively small training time cost, we achieved the best performance evaluation on Inception ResNet v2 by using data augmentation with image cropping. Fine-tuning demonstrates such an efficient and effective approach for learning DCNNs when data are scarce (Sermanet et al., 2013; Wang et al., 2017; Too et al., 2019). We thus conducted several experiments at a low initial learning rate (commonly $10^{-3}$ to $10^{-6}$) to unfreeze several or all the layers to investigate the architecture performance. To a certain extent, as the number of activated posterior layers increased, the test accuracy likewise increased (Fig. 7). Activation of shallow convolutional layers did not improve the performances of all four architectures. Only Inception ResNet v2 obtained the highest test accuracy by unfreezing all the layers (with 40 training epochs). The shallow convolutional layers that were trained on the larger ImageNet dataset exhibited high performance in detecting low-level features (such as brightness, edges, and curves in Fig. 3) on our dataset. Hence, our results underscore why fine-tuning has been used in many domains (Hentschel et al., 2016; Yin et al., 2017; Kaya et al., 2019).

The DCNNs demonstrated distinctive identification performances on different types of fossils, partially resulting from the intrinsic features of the fossil groups. The fossil or mineral groups have characteristically unique features (such as morphologies and structures) compared to other fossils that could be easily captured by the networks, such as mineral identification (pyrite and dolomite) compared to fossil groups. Here, we implemented t-distributed stochastic neighbor embedding (t-SNE) (Maaten and Hinton, 2008) to visualize the high-dimensional features extracted by Inception ResNet v2 (Fig. 9). We plotted 500 images (with an accuracy of 0.94) from the test dataset; among them, 1536 features were extracted from each image after the last Inception ResNet v2 convolutional layer.

By visualizing the extracted features from the DCNN architecture and the confusion matrix (Fig. 8), we noted that, although the machine classifier presented well-identified fossil groups with high generalization ability, it demonstrated ambiguous identification in four major categories: (1) algae versus foraminifer and sponge; (2) bivalve, brachiopod, and ostracod; (3) oncolite versus ooid and *Tubiphytes*; and (4) gastropod versus foraminifer and cephalopod. On the one hand, the misidentification between different classes with similar morphological features could imply the inherent difficulty that artificial neural networks have for classes in categories 2 and 3. Images from thin sections could contain random directions of the fossils, which leads to more fickle characteristic for each class. Although our data contains both greyscale and color images, the machine classifier demonstrates barely any difference of the performance applied to these images. We therefore transferred the greyscale images into "false" color matrixes for the input of the CNN architecture, and the neutral networks could capture the features when we provide enough training examples. Background noise also could influence the performance of the algorithms. In the 22 classes we include here, some of the groups are composed of plain morphologies, such as bivalve, ostracod, calcisphere, and calpionellid. The architectures could misidentify when the algorithms capture the background noise rather the true features of the fossils. Collecting more data and with appropriate image pre-processing methods could partially compensate for these deficiencies (Russakovsky et al., 2015). On the other hand, the composition of the dataset affects the performance of the machine classifier. In our collection, although more than two-thirds of our data was from the literature, several fossil groups remained insufficient, such as brachiopod, cephalopod, gastropod, radiolarian, and *Tubiphytes*. Therefore, we supplemented it with our materials from the Permian to Triassic in South China. Nevertheless, it was still difficult for us to distribute our data relatively evenly spread in high taxonomic groups for some classes (i.e., within-class imbalanced data), considering the wide morphological variation between high taxonomic groups in one fossil group, such as algae, foraminifers, and sponges. This deficiency resulted in identification biases within some classes, such as a higher accuracy on green algae compared to red algae (both identified as algae), a higher accuracy on crinoids compared to echinoids (both identified as echinoderm), and a higher accuracy on spiral-shelled cephalopods compared to straight-shelled cephalopods. Another within-dataset imbalanced situation is that the proportion of data from a few sources in the database is significantly higher than other sources. This could elevate the performance of the models during the experimental stage due to the higher similarities in the matrixes, background textures, and taphonomic regimes, but poorly perform in the real open database.

Hsiang et al. (2019) reported that a single class with a higher representation (abundance) leads to identification bias, which also results in decreased accuracy for undersampled foraminiferal species. There are several methods to partially compensate for imbalanced data (Chawla et al., 2002; Batista et al., 2004; He and Garcia, 2009; Krawczyk, 2016). The sampling method is the most common method by random oversampling of classes with minority data and undersampling the classes with majority data. Other methods, such as cost-sensitive methods, kernel-based methods, and active learning methods, are also frequently implemented to improve imbalanced datasets (He and Garcia, 2009). In our case, we removed some data from the classes (i.e., radiolarian, coral, and foraminifer) that had higher quantities to prevent highly skewed datasets.

Compared with human identification, the machine classifier works without prior experience and conducts identification based on the recognized image characteristics from the training phase. This implies that the CNN architecture can only identify the classes that have been previously trained. Any new class of images will still generate a probability vector belonging to those in classes that have been trained, as in the experimental studies of Pires de Lima et al. (2020a). Moreover, the machine classifier has the potential advantage of having high accuracy, reproducibility, and bias avoidance (Hsiang et al., 2019). In practice, the human classifier is common with accurate identification of several common taxa (having a higher occurrence) and poor identification of rare species. Even the accuracies of human experts are highly dependent on individual performance, often in unpredictable ways (Hsiang et al., 2019). Austen et al. (2018) conducted an investigation in which 17 experts were invited to identify four newt species from websites. The researchers found that additional years of experience did not improve the experts' identification performances; rather, increased expertise could result in a participant being more cautious in performing identification (Austen et al., 2016). Similarly, Al-Sabouni et al. (2018) determined that the length of an expert's experience did not correlate with higher identification accuracy for planktonic foraminifers. Meanwhile, taxonomists self-trained by reading books were more likely to exhibit lower accuracies with more divergent opinions compared to the community consensus. On the other hand, some of the species may only be reliably identified by a few core experts in the field but commonly misidentified by most practitioners (Hsiang et al., 2019). For example, Hsiang et al. (2019) noticed that a majority of images classified as *Globigerina bulloides* should actually be identified as *Globigerina falconensis*. The latter contains much rarer abundance and is unfamiliar even for some expert taxonomists. However, the performance of machine classifier mainly depends on inherent properties such as algorithms and original dataset. Another difference between human experts and machine classifiers is that human classifiers tend to be more phylogenetically conservative in their mistakes (Hsiang et al., 2019), i.e., the misidentified images commonly occur at the correct genus level. The machine classifier has mistaken identifications that occur in morphological similarity or even more random in all classes. In summary, the machine classifier now can reproduce instantly and objectively some recognition objectives that humans have already achieved. However, the means of addressing these results still depends on human experts (MacLeod et al., 2010). The CNN architectures also suffered from imbalanced data as discussed above, but less impacted compared with human classifiers (Hsiang et al., 2019). Hsiang et al. (2019) suggested that larger and unskewed samples can lead to more robust models, which can in turn result in a higher quality of machine accuracies that is comparable to (or may even surpass) human classifiers.

**5.2 Machine learning in petrography and paleontology**

Artificial neural networks in the microfacies classification that is based on petrographic thin sections have provided performances comparable to that of a petrographer. Previous studies have employed microfacies classification in both clastic sediments and carbonate rocks (Marmo et al., 2005; Pires de Lima et al., 2020a). Pires de Lima et al. (2020a) used the DCNN fine-tuning method to distinguish five similar siltstones, including argillaceous, bioturbated, and porous calcareous siltstone. Although their model can adapt DCNNs to achieve low error levels (<5%), their data were constrained by numerous sub-images, which were generated from a larger size of the original image. This led to the CNN classification having a higher accuracy for the particular sub-image during the analysis, whereas it engendered a lower generalization ability on other new images (Pires de Lima et al., 2020a). Another notable issue is that some of their training datasets contained images that were stained for specific minerals, such as calcite or dolomite identification; however, normally public data were missing those stained features that generalization ability would greatly reduce (Pires de Lima et al., 2020a). In further research, scholars can implement machine learning in semi-quantitative analysis of abstract features, such as permeability, porosity, and sorting levels of rock particles. This approach is somewhat subjective for human classifiers, especially for novices. These comprehensive structures can be divided into different sorting levels, such as very well sorted, moderately sorted, and poorly sorted (Shu et al., 2018), and different porosity levels, such as low, medium, and high porosity (Duarte-Coronado et al., 2019). In this circumstance, the petrographer can discretely and objectively label the porosity or sorting level of the training dataset to reduce interpretation bias. This can engender another significant issue with respect to biases due to the image. Thus, when addressing thin section heterogeneities, researchers should consider using the appropriate scale or size of the images for machine learning. The pre-processing used for the image cropping and scale resizing methods discussed previously is also worthy of caution. An example was demonstrated by Pires de Lima et al. (2020a), who found that the identification of level of bioturbations in thin sections were influenced by the scale of training images. Sometimes the bioturbation evidence is obscured when cropping the thin section images into smaller 10× magnification images.

Compared with previous studies on petrographic microfacies classifications, the present study is the first to focus on the identification of 18 common fossil group fragments which collected in thin sections from the Phanerozoic Eon. More than two-thirds of the data is from the literature, which provides our models with excellent generalization ability for both newly acquired data and our own samples. High taxonomic fossils groups (or even genera to species) identification in carbonate skeletons is essential for sedimentary environment analyses. In fact, our method is a semi-automatic identification method, which requires the preparation of a single fossil image for prediction. Subsequent research can be carried out from a real-time object detection task to implement the identification of multiple classes and fossils from a single microfacies image. Ultimately, fully automated carbonate petrography could be used to identify the fossils of the images captured from the camera in the microscope.

Although the earliest artificial intelligence techniques in biometric identification research have been employed since the 1970s, only a minimal amount of exploratory research has been conducted in the last several decades, mostly on invertebrates and microfossils (Pankhurst, 1974; MacLeod, 2007). Considering the costs of time, human resources, and financial resources, only a few segments of research can develop into continuous achievements to create useful popularization tools for biologists or paleontologists (MacLeod et al., 2010). Nonetheless, benefits have been provided by the rapid development of computer science in the past decade (Krizhevsky et al., 2012; Chetlur et al., 2014; Schmidhuber, 2015; Abadi et al., 2016), and many geological datasets are being digitized and standardized for sharing with peers, such as through publications and online databases (Hsiang et al., 2019). By collecting both online and offline geological big data and applying them for artificial intelligence use, human experts can be alleviated of the tedium of conducting routine identifications. For instance, we are striving to implement the microfacies automatic identification model that we trained in the present study in an online database for all petrographers and students. This will provide a convenient and efficient means for performing fossil identification during microfacies analysis.

# Conclusions

In this study, a deep learning algorithm that uses digital images to automatically classify fossil and abiotic grains from thin sections was implemented. Four classic DCNN architectures were trained on 30,815 images, which included 22 classes. All architectures obtained higher than 0.9 accuracies. Among them, the optimal model was demonstrated in the Inception ResNet v2 architecture. It exhibited 95% of the top one and 99% of the top three test accuracies. By using public data and our materials for model training, the architectures were more robust and had an improved generalization ability. Transfer learning is a powerful technology for implementing complex DCNN architectures in a relatively small dataset. The machine classifier performance was affected by data quality, data augmentation, and hyper-parameters; however, it showed the advantages of rapid identification, reproducibility, and bias avoidance compared with human classifiers. Automatic identification in petrography and paleontology based on DCNNs can provide geologists with an alternative and convenient method for routine and labor-intensive identification tasks.

# Acknowledgements

We thank Yongbiao Wang, Jing Chen, and Zheng Meng for supplying data, Minghui Wu, Zhuorong Li and Baichuan Jin for help on algorithm improvements, Shouyi Jiang for supporting hardware equipment in the early stage. We would also like to thank two anonymous reviewers for their insightful comments. This study is supported by the National Natural Science Foundation of China (41821001), the State Key R&D Project of China (2016YFA0601100), and Strategic Priority

**Figure and table captions**

**Fig. 1.** Example illustrations of each class in our dataset. Images of algae, brachiopod, bryozoan, gastropod, pyrite, and radiolarian are our own materials. Other photos are from the literature, including bivalve (Grosjean et al., 2018), calcimicrobe (Scholle and Ulmer-Scholle, 2003), calcisphere (Kröger et al., 2020), calpionellid (Akgümüş, 2019), cephalopod (Lakew, 1990), coral (McLean, 2005), dolomite (Franchi et al., 2016), echinoderm (Ghavidel-Syooki, 2017), foraminifer (Sartorio and Venturini, 1988), oncolite (Salama et al., 2014), ooid (Schlagintweit, 2008), ostracod (Donatelli and Tramontana, 2014), sponge and stromatoporoid (Flügel, 2010), stromatolite (Martin-Bello et al., 2019), and *Tubiphytes* (Senowbari-Daryan, 2013). Scale(s) for calpionellid is 0.05 mm, for calcisphere dolomite, and radiolarian are 0.1 mm, and for the others are 0.5 mm.

**Fig. 2.** Schematic of a convolutional neural network and its procedures. (A) A typical architecture of Visual Geometry Group (VGG)-16 network (Simonyan and Zisserman, 2014). It contains 13 convolutional layers, i.e., the blue rectangles; 5 max-pooling layers, i.e., the red rectangles; and three fully connected layers, i.e., purple rectangles. The last green rectangle is the softmax layer, which is the final output classes and generates the probability vector. (B) Schematic of a convolutional operation. Each image can be represented by a pixel matrix, its shape is height × width × channels. Here, we demonstrated a 5 px × 5 px matrix as the input image, a 2 × 2 matrix of filter (yellow rectangle) will scroll (from the left-to-right, top-to-bottom, stride = 1) in the input matrix and computes the inner product with the overlap matrix (brown rectangle), the results will generate a new matrix, also called feature map or activation map. (C) A max-pooling operation with stride equals 2. The adjacent pixels have similar structures, pooling layer aims to downsampling. (D) Fully connected layers in the top of the VGG-16. Every neuron in one layer will connect with every neuron in the next layer. (E) Schematic of an artificial neuron.

**Fig. 3.** Visualization of the feature maps and heatmaps from different VGG-16 convolutional layers. The heatmaps are fusion maps after the conv5 layers. The shallower layers detected low-level features such as brightness, edges, and curves, while the deeper layers extract more abstract features. By stacking several convolutional and pooling layers, we could gradually encode higher-level feature representations (Gu et al., 2018).

**Fig. 4.** Confusion matrix and performance evaluation metrics calculated from it, modified from Fawcett (2006).

**Fig. 5.** Schematic representation of the other three DCNN architectures we used. (A) ResNet v1-152 architecture (He et al., 2016), (B) Inception v4 architecture (Szegedy et al., 2016), and (C) Inception ResNet v2 architecture (Szegedy et al., 2016). The ResNet block contains four types of residual blocks, each residual block stacks up its input (identity mapping) and output (residual mapping) to be the input of the next layer. The identity mappings can prevent the vanishing gradient issue. Each ResNet block implemented several times to decrease dimensions of the feature maps and increase the depth of the network. The stem module is a pre-processing step before the feature maps inputs to the Inception (ResNet) module, it contains two branches: convolutional and pooling branches. By combining two branches to significantly increase the channels for better feature extraction. The Inception module contains three types of Inception blocks, each module embeds a multiscale feature extractor, 1× 1, 3 × 3, and 5 × 5 etc. convolutional layers. Each module was repeated several times to increase the depth of the network. The Inception ResNet module includes three kinds of modules, each module combined the characteristic of ResNet block and Inception module. The Inception ResNet modules decrease the channels and increase the depth of the architecture compared with modules from Inception v4. The Reduction module aims to reduce the dimensions to half, and increase the channels to make up for dimension decrease, it contains three or four branches and composed of one max pooling branch and two or three convolutional branches.

**Fig. 6.** Curves showing the evaluation of the (A) training loss, (B) training accuracy, (C) validation loss, and (D) validation accuracy with epoch for four DCNNs during the training processes. The hyper-parameter of these four DCNNs come from analysis number #1, #7, #13, and #19 in Table 2.

**Fig. 7.** Curves showing the evaluation of the (A) training loss, (B) training accuracy, (C) validation loss, and (D) validation accuracy with epoch during the training processes for Inception ResNet v2 with different hyper-parameter. The hyper-parameter of these four training models come from the analysis number #20–#23 in Table 2.

**Fig. 8.** Confusion matrix demonstrates true and predicted labels from the validation and test datasets, which tested on Inception ResNet v2 architecture.

**Fig. 9.** Visualization of the feature spatial distribution extracted by Inception ResNet v2 architecture from the 500 images (with 94.2% of accuracy) in the test dataset using t-SNE. There are 1536 features extracted from each test image after the last Inception ResNet v2 convolutional layer. Each number and colour quadrangle represent a class, the class order corresponding to alphabetical order

in Table 1.

**Table 1.** Quantities of the training set, validation set, and test set for each fossil and abiotic grain group.

**Table 2.** The analysis results of four DCNN architectures. The network: 1 = VGG-16; 2 = ResNet v1-152; 3 = Inception v4; 4 = Inception ResNet v2. Load weights are load the pre-trained parameters from the ImageNet for variable initialize. Frozen layers are un-trainable layers. In the train layers column, all layers suggest that all layers of the architecture are trainable; half layers mean we trained the posterior half part of the architecture: for VGG-16 we trained conv4, conv5, cf6, fc7, and fc8 (Fig. 1A); for ResNet v1-152 we trained Block3, Block4, and the final layer (Fig. 5A); for Inception v4 we trained inception-B, Inception-C, and the final layer (Fig. 5B); for Inception ResNet v2 we trained Inception ResNet-B, Inception ResNet-C, and the final layer (Fig. 5C). Drop out range from 0−1, where 0 stands for probabilistically removing or "dropping out" all the inputs to the next layer, 1 stands for no dropping out. Decay step stands for each corresponding iteration, the learning rate will multiply the decay rate. Batch nor. = Batch normalization. This function aims to scale and standardize the input training data before feeding it to a model for training and benefit to model convergence. Num aug. = Number of methods for data augmentation. For VGG and ResNet, 3 means no crop of training images (i.e., methods (1), (3), and (5) in section 3.3); 4 means randomly crop training images; For Inception networks, 4 means no crop of training images (i.e., methods (1), (3), (5), and (6) in section 3.3), and 5 means randomly crop training images. Optimizer is the algorithm or method used to optimize the performance of the neural networks by providing the appropriate optimization direction.

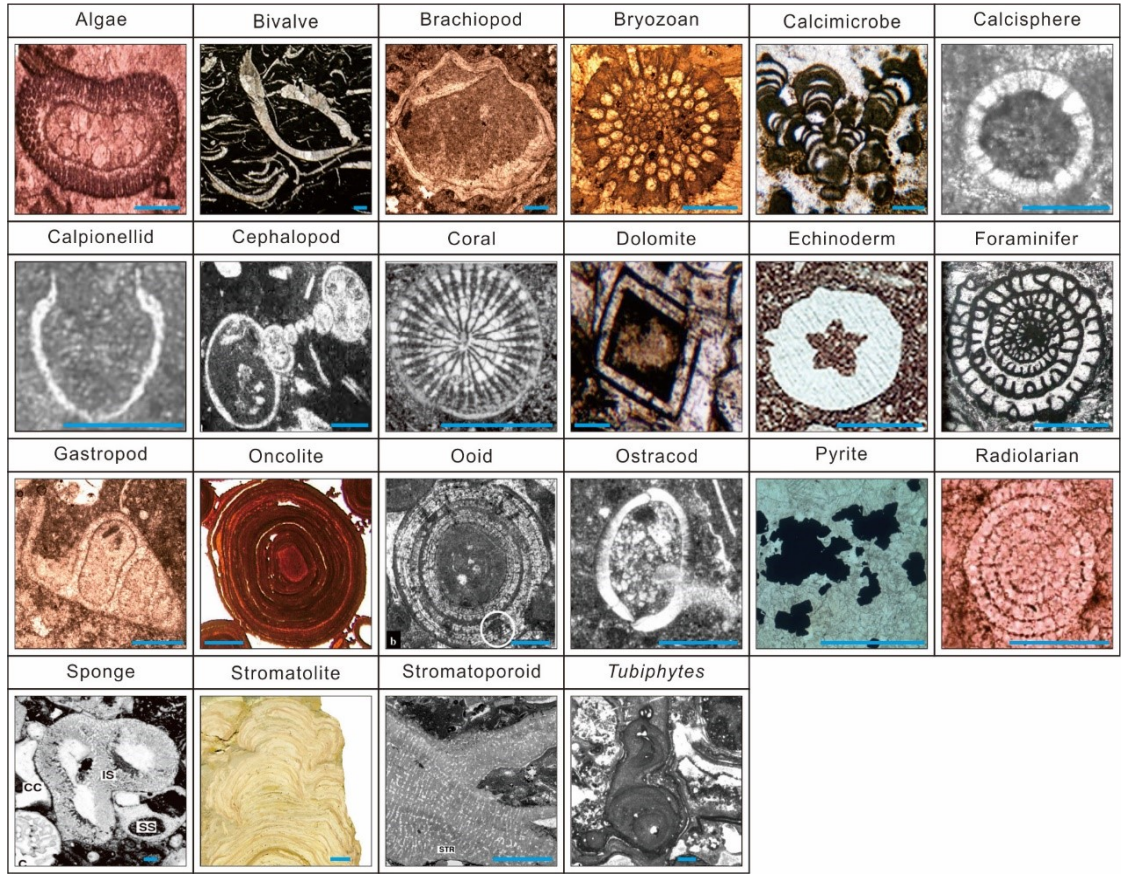

**Fig. 1**

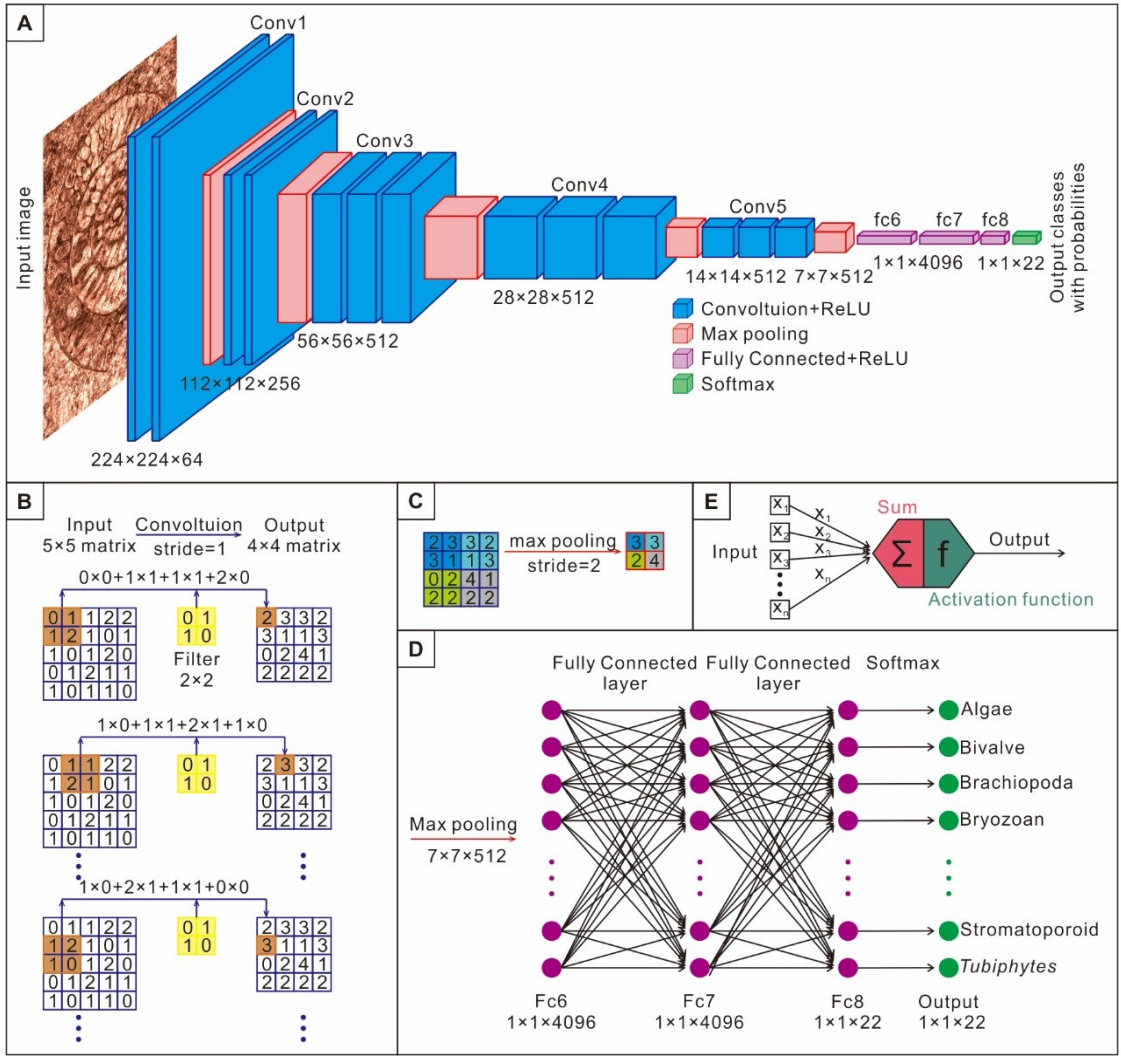

**Fig. 2**

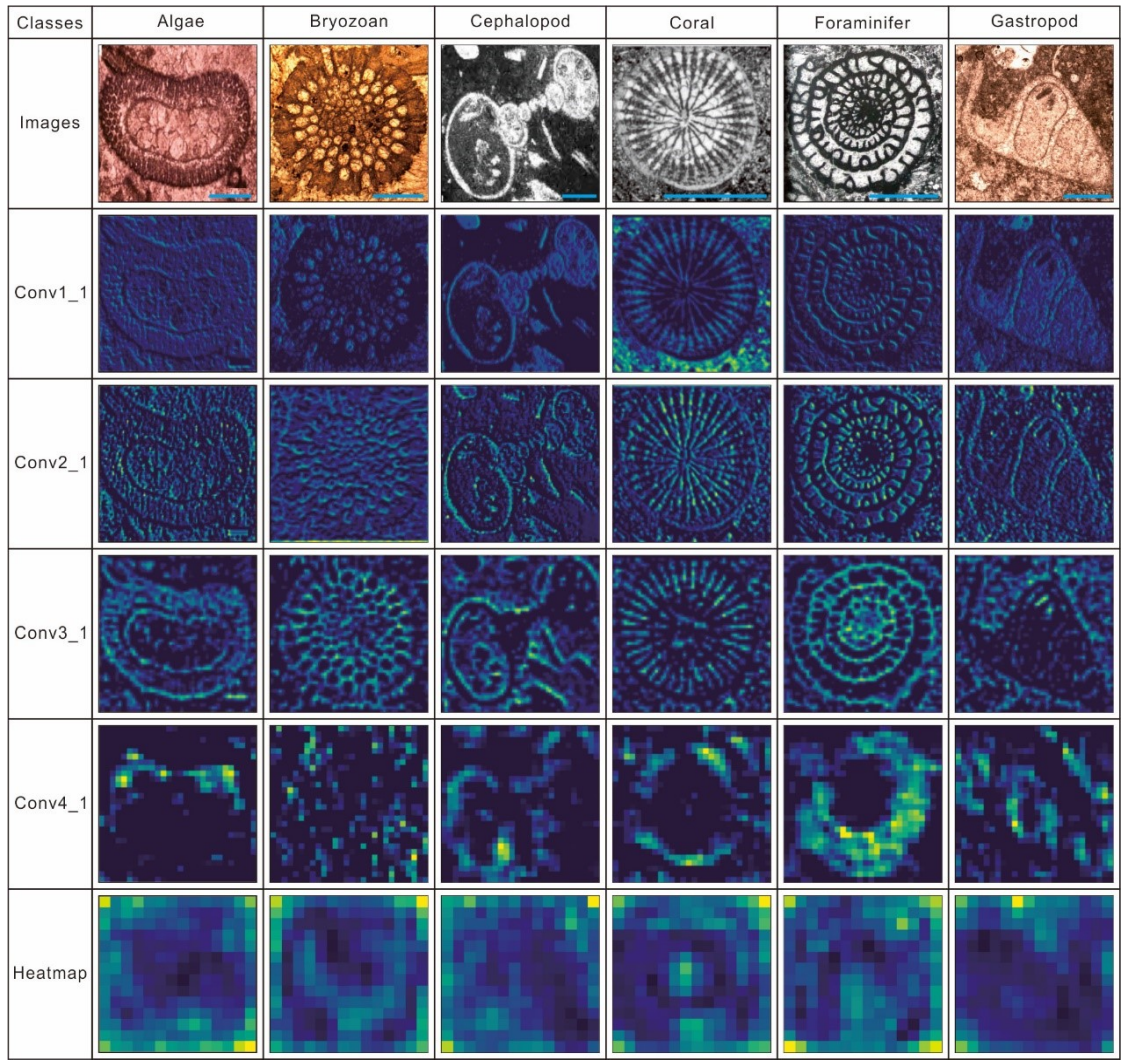

**Fig. 3**

| Confusion matrix | | True labels | | Evaluation |
|---|---|---|---|---|
| | | Positive | Negative | |
| Predicted labels | Positive | True Positive | False Positive | Precision $\frac{TP}{(TP+FP)}$ |
| | Negative | False Negative | True Negative | F1 score $\frac{2*Prescision*Recall}{(Prescision+Recall)}$ |
| Evaluation | | Recall $\frac{TP}{(TP+FN)}$ | | Accuracy $\frac{TP+TN}{(TP+TN+FP+FN)}$ |

**Fig. 4**

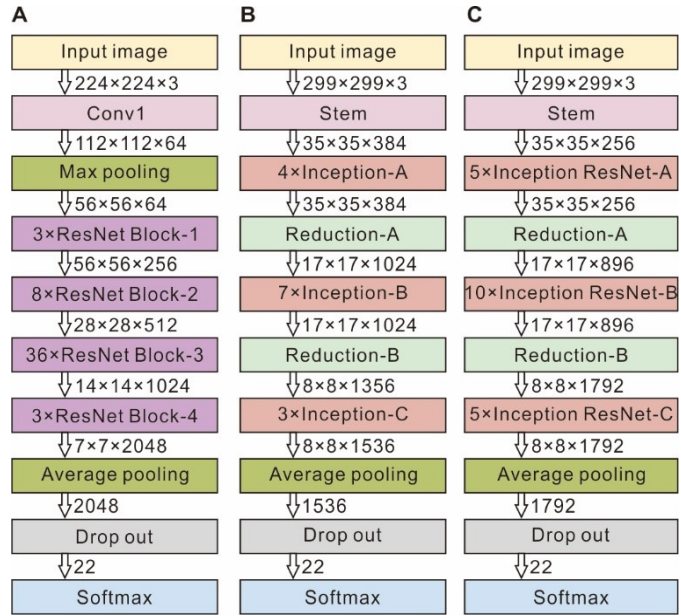

**Fig. 5**

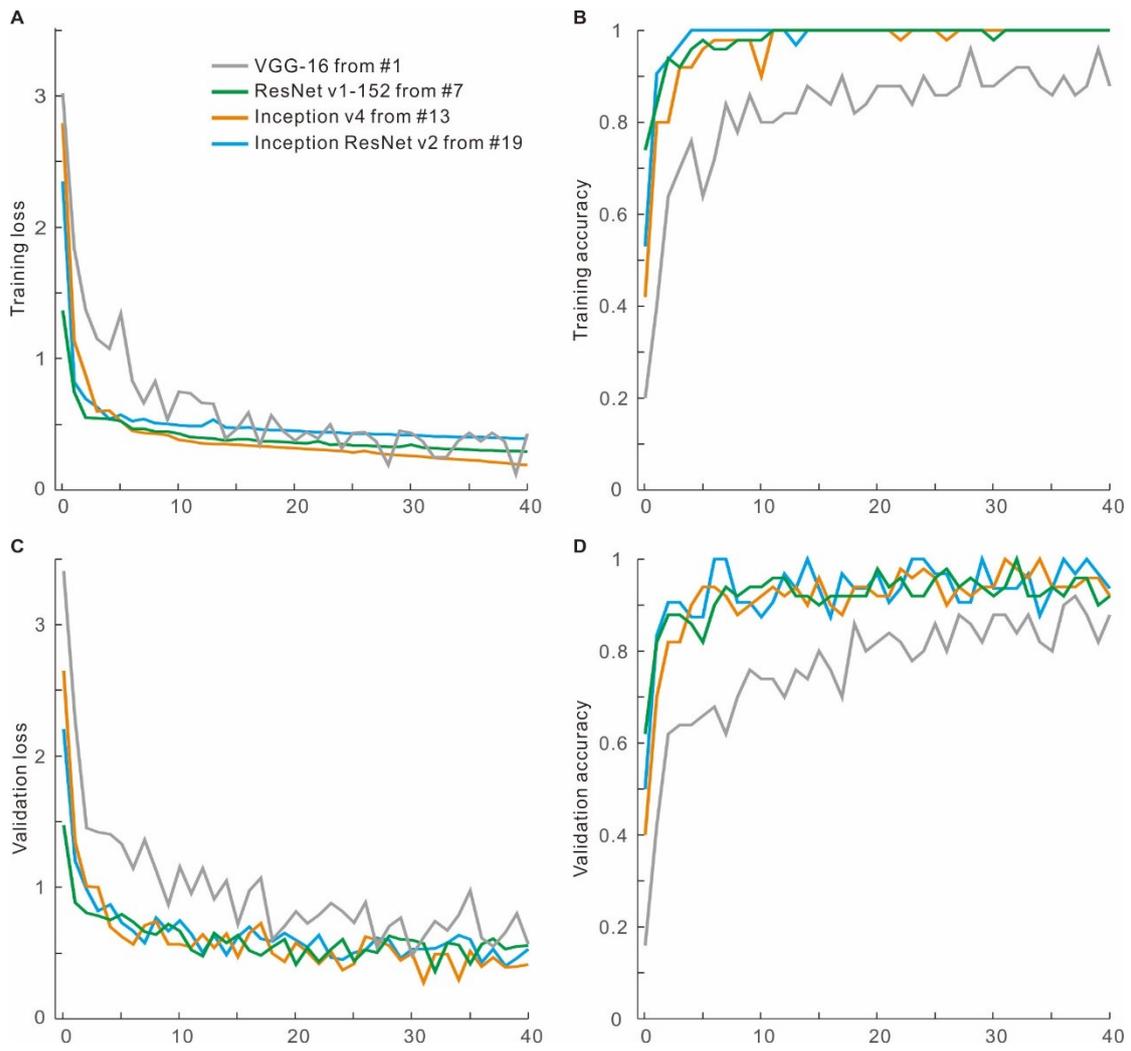

**Fig. 6**

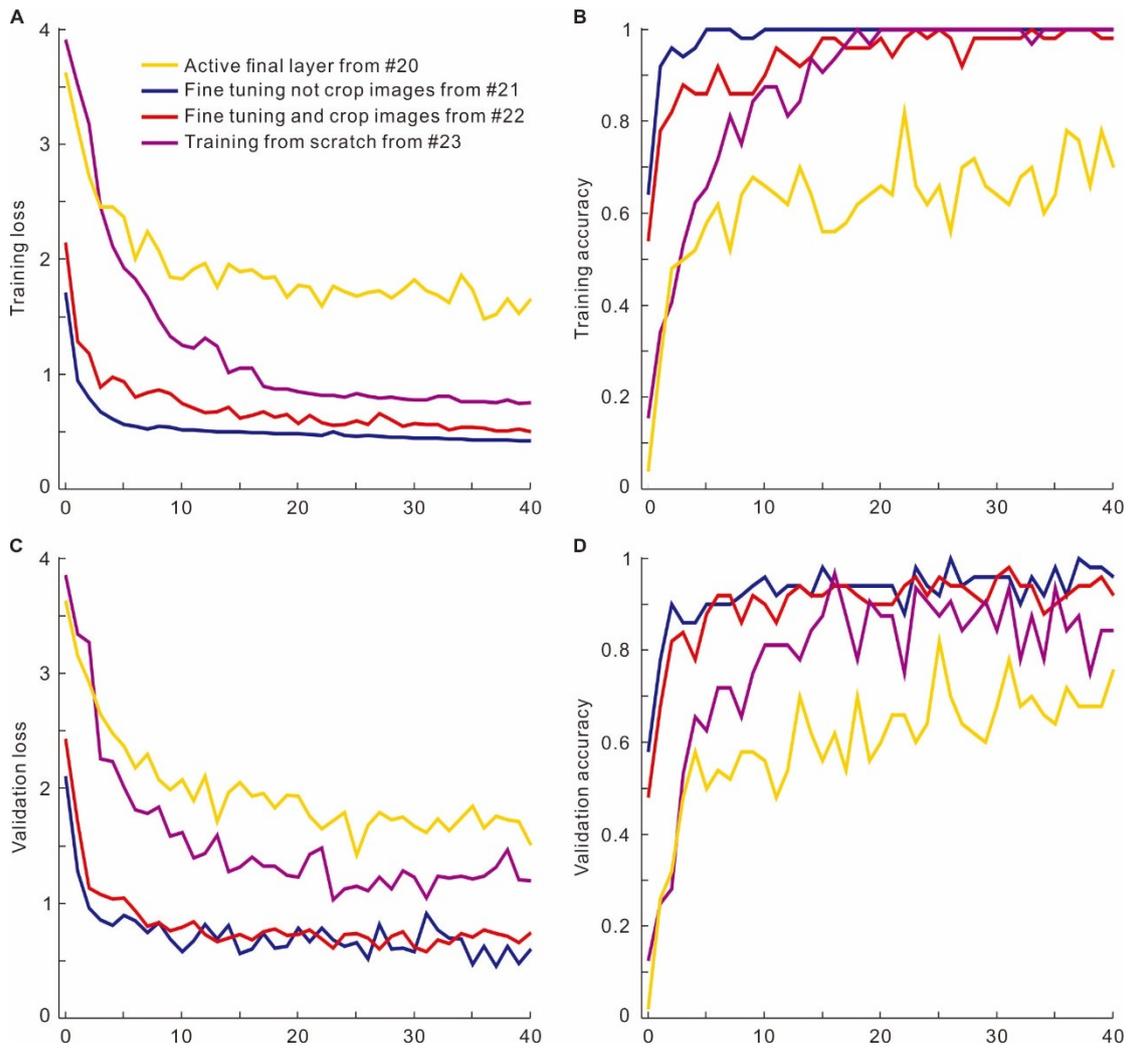

Fig. 7

| True class | Total number | Precision | Recall | F1 score | Algae | Bivalve | Brachiopod | Bryozoan | Calcimicrobe | Calcisphere | Calpionellid | Cephalopod | Coral | Dolomite | Echinoderm | Foraminifer | Gastropod | Oncolite | Ooid | Ostracod | Pyrite | Radiolarian | Sponge | Stromatolite | Stromatoporoid | Tubiphytes |
|---|---|---|---|---|---|---|---|---|---|---|---|---|---|---|---|---|---|---|---|---|---|---|---|---|---|---|
| Algae | 241 | 0.91 | 0.86 | 0.89 | 0.91 | 0.00 | 0.00 | 0.01 | 0.02 | 0.00 | 0.00 | 0.00 | 0.00 | 0.00 | 0.00 | 0.02 | 0.00 | 0.00 | 0.00 | 0.00 | 0.00 | 0.00 | 0.02 | 0.00 | 0.00 | 0.00 |
| Bivalve | 240 | 0.89 | 0.86 | 0.87 | 0.00 | 0.89 | 0.05 | 0.00 | 0.00 | 0.00 | 0.00 | 0.00 | 0.00 | 0.00 | 0.01 | 0.00 | 0.01 | 0.00 | 0.02 | 0.00 | 0.00 | 0.00 | 0.00 | 0.00 | 0.00 | 0.01 |
| Brachiopod | 246 | 0.89 | 0.85 | 0.87 | 0.00 | 0.09 | 0.89 | 0.00 | 0.00 | 0.00 | 0.00 | 0.00 | 0.00 | 0.00 | 0.00 | 0.00 | 0.00 | 0.00 | 0.02 | 0.00 | 0.00 | 0.00 | 0.00 | 0.00 | 0.00 | 0.00 |
| Bryozoan | 293 | 0.92 | 0.89 | 0.91 | 0.02 | 0.00 | 0.00 | 0.92 | 0.00 | 0.00 | 0.00 | 0.00 | 0.02 | 0.00 | 0.00 | 0.01 | 0.00 | 0.00 | 0.00 | 0.00 | 0.00 | 0.02 | 0.00 | 0.00 | 0.00 | 0.00 |
| Calcimicrobe | 250 | 0.92 | 0.91 | 0.92 | 0.01 | 0.00 | 0.00 | 0.02 | 0.92 | 0.00 | 0.00 | 0.00 | 0.01 | 0.00 | 0.00 | 0.00 | 0.00 | 0.00 | 0.00 | 0.00 | 0.00 | 0.00 | 0.02 | 0.00 | 0.01 | 0.01 |
| Calcisphere | 249 | 0.91 | 0.95 | 0.93 | 0.00 | 0.00 | 0.00 | 0.00 | 0.00 | 0.91 | 0.00 | 0.00 | 0.00 | 0.00 | 0.00 | 0.00 | 0.02 | 0.00 | 0.01 | 0.02 | 0.00 | 0.02 | 0.00 | 0.00 | 0.00 | 0.00 |
| Calpionellid | 295 | 0.99 | 0.99 | 0.99 | 0.00 | 0.00 | 0.00 | 0.00 | 0.00 | 0.00 | 0.01 | 0.99 | 0.00 | 0.00 | 0.00 | 0.00 | 0.00 | 0.00 | 0.00 | 0.00 | 0.00 | 0.00 | 0.00 | 0.00 | 0.00 | 0.00 |
| Cephalopod | 256 | 0.97 | 0.94 | 0.95 | 0.00 | 0.00 | 0.00 | 0.00 | 0.00 | 0.00 | 0.00 | 0.00 | 0.97 | 0.00 | 0.00 | 0.00 | 0.00 | 0.03 | 0.00 | 0.00 | 0.00 | 0.00 | 0.00 | 0.00 | 0.00 | 0.00 |
| Coral | 354 | 0.94 | 0.97 | 0.96 | 0.03 | 0.00 | 0.00 | 0.02 | 0.00 | 0.00 | 0.00 | 0.00 | 0.00 | 0.94 | 0.00 | 0.00 | 0.00 | 0.00 | 0.00 | 0.00 | 0.00 | 0.00 | 0.00 | 0.00 | 0.00 | 0.00 |
| Dolomite | 260 | 1.00 | 0.99 | 0.99 | 0.00 | 0.00 | 0.00 | 0.00 | 0.00 | 0.00 | 0.00 | 0.00 | 0.00 | 0.00 | 1.00 | 0.00 | 0.00 | 0.00 | 0.00 | 0.00 | 0.00 | 0.00 | 0.00 | 0.00 | 0.00 | 0.00 |
| Echinoderm | 318 | 0.95 | 0.95 | 0.95 | 0.01 | 0.00 | 0.01 | 0.01 | 0.00 | 0.00 | 0.00 | 0.00 | 0.00 | 0.00 | 0.00 | 0.95 | 0.01 | 0.00 | 0.00 | 0.00 | 0.00 | 0.00 | 0.00 | 0.00 | 0.00 | 0.00 |
| Foraminifer | 291 | 0.95 | 0.92 | 0.93 | 0.01 | 0.00 | 0.00 | 0.00 | 0.01 | 0.01 | 0.00 | 0.00 | 0.00 | 0.00 | 0.00 | 0.00 | 0.95 | 0.00 | 0.00 | 0.00 | 0.00 | 0.00 | 0.00 | 0.01 | 0.01 | 0.00 | 0.00 |
| Gastropod | 294 | 0.89 | 0.92 | 0.90 | 0.00 | 0.01 | 0.00 | 0.00 | 0.01 | 0.00 | 0.00 | 0.00 | 0.04 | 0.00 | 0.00 | 0.00 | 0.01 | 0.89 | 0.01 | 0.00 | 0.01 | 0.00 | 0.00 | 0.02 | 0.00 | 0.00 | 0.00 |
| Oncolite | 313 | 0.85 | 0.89 | 0.87 | 0.01 | 0.01 | 0.00 | 0.01 | 0.01 | 0.00 | 0.00 | 0.00 | 0.00 | 0.00 | 0.00 | 0.00 | 0.00 | 0.01 | 0.85 | 0.05 | 0.00 | 0.00 | 0.00 | 0.00 | 0.01 | 0.00 | 0.04 |
| Ooid | 295 | 0.94 | 0.92 | 0.93 | 0.01 | 0.00 | 0.00 | 0.00 | 0.00 | 0.00 | 0.00 | 0.00 | 0.00 | 0.00 | 0.00 | 0.00 | 0.00 | 0.03 | 0.94 | 0.01 | 0.00 | 0.00 | 0.00 | 0.00 | 0.00 | 0.00 |
| Ostracod | 336 | 0.88 | 0.93 | 0.90 | 0.00 | 0.01 | 0.06 | 0.00 | 0.00 | 0.01 | 0.01 | 0.00 | 0.00 | 0.00 | 0.00 | 0.01 | 0.01 | 0.00 | 0.00 | 0.88 | 0.00 | 0.00 | 0.00 | 0.00 | 0.00 | 0.00 |
| Pyrite | 248 | 0.99 | 1.00 | 0.99 | 0.00 | 0.00 | 0.00 | 0.00 | 0.00 | 0.00 | 0.00 | 0.00 | 0.00 | 0.00 | 0.00 | 0.00 | 0.00 | 0.00 | 0.00 | 0.00 | 0.99 | 0.00 | 0.00 | 0.00 | 0.00 | 0.00 |
| Radiolarian | 319 | 0.99 | 0.97 | 0.98 | 0.00 | 0.00 | 0.00 | 0.00 | 0.00 | 0.00 | 0.00 | 0.00 | 0.00 | 0.00 | 0.00 | 0.00 | 0.00 | 0.00 | 0.00 | 0.00 | 0.00 | 0.99 | 0.00 | 0.00 | 0.00 | 0.00 |
| Sponge | 296 | 0.92 | 0.87 | 0.90 | 0.01 | 0.00 | 0.00 | 0.01 | 0.01 | 0.00 | 0.00 | 0.00 | 0.01 | 0.00 | 0.01 | 0.00 | 0.01 | 0.00 | 0.00 | 0.00 | 0.00 | 0.01 | 0.92 | 0.00 | 0.01 | 0.00 |
| Stromatolite | 269 | 0.89 | 0.98 | 0.93 | 0.01 | 0.01 | 0.00 | 0.01 | 0.02 | 0.00 | 0.00 | 0.00 | 0.00 | 0.00 | 0.00 | 0.00 | 0.00 | 0.02 | 0.01 | 0.00 | 0.00 | 0.00 | 0.01 | 0.89 | 0.01 | 0.00 |
| Stromatoporoid | 262 | 0.95 | 0.97 | 0.96 | 0.00 | 0.01 | 0.00 | 0.00 | 0.00 | 0.00 | 0.00 | 0.00 | 0.01 | 0.00 | 0.00 | 0.00 | 0.00 | 0.00 | 0.00 | 0.00 | 0.00 | 0.01 | 0.00 | 0.95 | 0.00 |
| *Tubiphytes* | 275 | 0.93 | 0.93 | 0.93 | 0.01 | 0.00 | 0.00 | 0.00 | 0.00 | 0.00 | 0.00 | 0.00 | 0.00 | 0.01 | 0.00 | 0.00 | 0.04 | 0.00 | 0.00 | 0.00 | 0.00 | 0.01 | 0.00 | 0.00 | 0.93 |

True classes of validation and test images (rows); Predicted classes of validation and test images (columns).

**Fig. 8**

Fig. 9

| Order | Classes | Training set (80%) | Validation set (15%) | Test set (5%) | Total |
|---|---|---|---|---|---|
| 0 | Algae | 1037 | 195 | 64 | 1296 |
| 1 | Bivalve | 993 | 186 | 62 | 1241 |
| 2 | Brachiopod | 1011 | 189 | 63 | 1263 |
| 3 | Bryozoan | 1162 | 218 | 72 | 1452 |
| 4 | Calcimicrobe | 1036 | 194 | 64 | 1294 |
| 5 | Calcisphere | 982 | 184 | 61 | 1227 |
| 6 | Calpionellid | 1129 | 212 | 70 | 1411 |
| 7 | Cephalopod | 1039 | 195 | 64 | 1298 |
| 8 | Coral | 1317 | 247 | 82 | 1646 |
| 9 | Dolomite | 1023 | 192 | 63 | 1278 |
| 10 | Echinoderm | 1260 | 237 | 78 | 1575 |
| 11 | Foraminifer | 1260 | 236 | 78 | 1574 |
| 12 | Gastropod | 1135 | 213 | 70 | 1418 |
| 13 | Oncolite | 1218 | 228 | 76 | 1522 |
| 14 | Ooid | 1168 | 219 | 73 | 1460 |
| 15 | Ostracod | 1288 | 242 | 80 | 1610 |
| 16 | Pyrite | 996 | 186 | 62 | 1244 |
| 17 | Radiolarian | 1259 | 236 | 78 | 1573 |
| 18 | Sponge | 1223 | 229 | 76 | 1528 |
| 19 | Stromatolite | 1025 | 192 | 64 | 1281 |
| 20 | Stromatoporoid | 998 | 187 | 62 | 1247 |
| 21 | *Tubiphytes* | 1102 | 207 | 68 | 1377 |
|  | Total | 24661 | 4624 | 1530 | 30815 |

**Table 1**

| Order | Network | Batch size | Load weights | Frozen layers | Train layers | Drop out | Start learning rate | Decay step | Decay rate | Batch nor. | Num aug. | Optimizer | Epoch ran | Max tra acc | Min tra loss | Max val acc | Min val loss | Max top 1 test acc | Max top 3 test acc |
|---|---|---|---|---|---|---|---|---|---|---|---|---|---|---|---|---|---|---|---|
| 1 | 1 | 50 | Yes | Yes | Half layers | 0.8 | 0.0001 | 400 | 0.96 | Yes | 3 | Adam | 50 | 0.96 | 0.12 | 0.84 | 0.52 | 0.91 | 0.98 |
| 2 | 1 | 50 | Yes | Yes | Half layers | 0.8 | 0.00001 | 1000 | 0.96 | Yes | 3 | Adam | 50 | 0.96 | 0.15 | 0.84 | 0.66 | 0.88 | 0.96 |
| 3 | 1 | 50 | Yes | No | All layers | 0.8 | 0.0001 | 400 | 0.96 | No | 4 | Adam | 50 | 1.00 | 0.43 | 0.94 | 0.61 | 0.88 | 0.97 |
| 4 | 1 | 100 | Yes | Yes | Half layers | 0.8 | 0.0001 | 400 | 0.96 | Yes | 3 | Adam | 40 | 1.00 | 0.00 | 0.91 | 0.48 | 0.86 | 0.97 |
| 5 | 1 | 50 | No | No | All layers | 0.8 | 0.0001 | No | | No | 3 | Adam | 60 | 1.00 | 0.55 | 0.72 | 1.65 | 0.59 | 0.71 |
| 6 | 1 | 50 | Yes | Yes | Half layers | 0.5 | 0.0001 | 400 | 0.96 | No | 3 | Adam | 50 | 1.00 | 0.78 | 0.96 | 0.91 | 0.91 | 0.98 |
| 7 | 2 | 50 | Yes | Yes | Half layers | 1.0 | 0.0001 | 400 | 0.96 | Yes | 3 | Adam | 40 | 1.00 | 0.29 | 1.00 | 0.36 | 0.93 | 0.99 |
| 8 | 2 | 50 | Yes | Yes | Half layers | 0.8 | 0.0001 | 400 | 0.96 | Yes | 4 | RMSP | 40 | 1.00 | 0.38 | 0.98 | 0.49 | 0.91 | 0.98 |
| 9 | 2 | 50 | Yes | Yes | All layers | 0.8 | 0.0001 | 400 | 0.96 | Yes | 3 | Adam | 40 | 1.00 | 0.27 | 0.98 | 0.35 | 0.91 | 0.99 |
| 10 | 2 | 50 | Yes | Yes | Half layers | 1.0 | 0.001 | 300 | 0.96 | Yes | 3 | Adam | 40 | 1.00 | 0.19 | 0.96 | 0.46 | 0.89 | 0.98 |
| 11 | 2 | 50 | No | No | All layers | 0.8 | 0.0001 | 500 | 0.96 | Yes | 3 | Adam | 60 | 1.00 | 2.39 | 0.82 | 3.50 | 0.68 | 0.86 |
| 12 | 2 | 64 | Yes | Yes | Half layers | 0.8 | 0.0001 | 400 | 0.96 | Yes | 3 | Adam | 40 | 1.00 | 0.10 | 0.95 | 0.34 | 0.90 | 0.97 |
| 13 | 3 | 50 | Yes | Yes | Half layers | 0.8 | 0.0001 | 400 | 0.96 | Yes | 4 | RMSP | 40 | 1.00 | 0.19 | 1.00 | 0.28 | 0.94 | 0.99 |
| 14 | 3 | 50 | Yes | No | All layers | 0.8 | 0.0001 | 400 | 0.96 | Yes | 4 | RMSP | 40 | 1.00 | 0.26 | 0.98 | 0.37 | 0.94 | 0.99 |
| 15 | 3 | 50 | Yes | Yes | Half layers | 0.5 | 0.0001 | 400 | 0.96 | Yes | 4 | RMSP | 40 | 1.00 | 0.31 | 0.98 | 0.44 | 0.93 | 0.99 |
| 16 | 3 | 50 | Yes | Yes | Half layers | 0.8 | 0.0001 | No | | Yes | 5 | RMSP | 40 | 1.00 | 0.52 | 0.96 | 0.54 | 0.89 | 0.98 |
| 17 | 3 | 32 | Yes | Yes | Half layers | 0.8 | 0.00001 | 1000 | 0.96 | Yes | 5 | RMSP | 40 | 0.94 | 0.77 | 0.94 | 0.60 | 0.82 | 0.95 |
| 18 | 3 | 50 | No | No | All layers | 0.8 | 0.0001 | 400 | 0.96 | Yes | 4 | RMSP | 60 | 1.00 | 1.12 | 0.90 | 1.75 | 0.73 | 0.90 |
| 19 | 4 | 32 | Yes | Yes | All layers | 0.8 | 0.0001 | 400 | 0.96 | Yes | 4 | Adam | 40 | 1.00 | 0.39 | 1.00 | 0.40 | 0.95 | 0.99 |
| 20 | 4 | 50 | Yes | Yes | Last layer | 0.8 | 0.0001 | 400 | 0.96 | Yes | 4 | Adam | 40 | 0.82 | 1.49 | 0.82 | 1.42 | 0.67 | 0.86 |
| 21 | 4 | 50 | Yes | Yes | Half layers | 0.8 | 0.0001 | 400 | 0.96 | Yes | 4 | Adam | 40 | 1.00 | 0.42 | 1.00 | 0.45 | 0.94 | 0.99 |
| 22 | 4 | 50 | Yes | Yes | Half layers | 0.8 | 0.0001 | 400 | 0.96 | Yes | 5 | Adam | 40 | 1.00 | 0.50 | 0.98 | 0.58 | 0.93 | 0.99 |
| 23 | 4 | 32 | No | No | All layers | 0.8 | 0.0001 | 400 | 0.96 | Yes | 4 | Adam | 60 | 1.00 | 0.71 | 0.97 | 0.83 | 0.82 | 0.95 |
| 24 | 4 | 100 | Yes | Yes | Half layers | 0.8 | 0.0001 | 400 | 0.96 | Yes | 4 | Adam | 40 | 1.00 | 0.00 | 0.96 | 0.08 | 0.94 | 0.99 |

**Table 2**